\documentclass[11pt,a4paper]{article}
\usepackage[hyperref]{acl2019}
\usepackage{times}
\usepackage{phonetic}
\usepackage{latexsym}
\usepackage{tikz}
\usetikzlibrary{arrows.meta,
                chains,
                positioning,
                shapes.misc}
                
\usepackage{url}

\aclfinalcopy 

\usepackage{xspace}
\usepackage[backgroundcolor=green!35]{todonotes}

\title{Semantic Change and Semantic Stability: Variation is Key}

\author{Claire L.\ Bowern \\
  Yale University / 370 Temple St \\
  New Haven, CT 06520 \\
  USA \\
  \texttt{claire.bowern@yale.edu} }

\date{\today}

\begin{document}
\maketitle

\begin{abstract}
I survey some recent approaches to studying change in the lexicon, particularly change in meaning across phylogenies. I briefly sketch an evolutionary approach to language change and point out some issues in recent approaches to studying semantic change that rely on temporally stratified word embeddings. I draw illustrations from lexical cognate models in Pama-Nyungan to identify meaning classes most appropriate for lexical phylogenetic inference, particularly highlighting the importance of variation in studying change over time.
\end{abstract}

\section{Introduction}
All aspects of all languages are changing all the time. And for most of human history, for most of the world's languages, this change is not recorded. Therefore, in order to understand language change adequately, we need methods which allow us to extrapolate back beyond what is identifiable in the written record, which is both shallow and geographically sparse. In this paper, I discuss how evolutionary approaches to language change allow the modeling of cognate evolution. I show how such models can be used to study semantic change at the macro-level, and finally how we can make use of existing data to refine meaning categories for use in inferring language splits. I focus on theoretical models of change.







I begin with a brief outline of contemporary language change, particularly as studied quantitatively (\citealt{bow18} provides more context). I then discuss issues of reconstructing meaning and identifying meaning change, before presenting two case studies: one on studying semantic change across a phylogeny, the other about identifying lexical stability.

\subsection{What is language change}\label{sec:keyqs}
Much contemporary work on historical linguistics aims to answer one or more of three key questions for the nature of language change:

\begin{enumerate}
    \item \emph{What} forms have changed?\vspace*{-4pt} 
    \item \emph{How} does change work?\vspace*{-4pt} 
    \item \emph{Why} does it work the way it does?\vspace*{-4pt}
\end{enumerate}

The first aspect of diachrony involves establishing the ``facts'': that is, identifying differences between languages at various stages of their history (or differences between related languages) and establishing which of those differences are due to \emph{change} in the system, and which are artefacts of data gathering or sampling. Research of this type includes how language informs our study of prehistory. Questions of this type include ``Where was the homeland of speakers of Proto-Pama-Nyungan?'' \cite{bouckaert2018} or ``What is the origin of  the Latin ablative case?''

The second question -- how does change work -- seeks to establish the general properties of change. These are ``mode and tempo'' type questions \citep{greenhill2010shape}, regarding which items in language change more rapidly than others, what features change into which others, and which features are stable across centuries and millennia. Work in this area include \citet{hamilton2016diachronic} on semantic change, \citet{wedelfunctional2013} on sound change, \citet{van2018diachrony} on change in argument structure, and indeed much  work of recent years \citep{bowernevans2014}.

The third question -- the \emph{why} of language change -- has received less attention. Until recently, it has been difficult to study changes at the scale necessary, and with the precision necessary, to do more than speculate. Moreover, the focus in historical linguistics on language-internal explanations has made it difficult to grapple with the obvious fact that languages change in large part because of the way people acquire and use them (see further \S\ref{seq:trad}). One example of modeling a `why' of change in meaning comes from \citet{ahern2017conflict}, which argues that one type of semantic change occurs because of psychological tendencies for interlocutors to assume exaggeration.


\begin{figure}[ht]
    \centering
\begin{tikzpicture}[
roundnode/.style={circle, draw=green!60, fill=green!5, very thick, minimum size=6mm}]

\node[]        (align1)       [] {};
\node[roundnode]        (what)       [right=of align1, align=center] {What};
\node[]        (align2)       [right=of what,align=center] {};
\node[roundnode]        (how)       [below=of align1,align=center] {How};
\node[roundnode]        (why)       [below=of align2,align=center] {Why};

\begin{scope}[>=Latex]
\draw[<->, thick] (what.west) -- (how.north);
\draw[<->, thick] (what.east) -- (why.north);
\draw[<->, thick] (how.east) -- (why.west);
\end{scope}
\end{tikzpicture}		
\centering
\caption{Key questions}\label{fig:keyqs}
\end{figure}
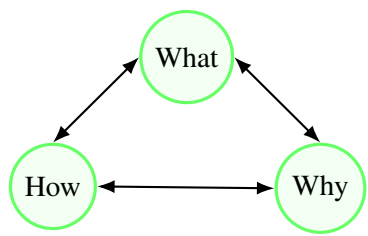

All of these questions are related to one another, and the answers to one inform the others. We cannot make plausible inferences about processes without a theory, any more than we can work on a theory of change without data to test it with. The \emph{what} provides us with observations; the \emph{why} provides us with a theory that explains those observations, and the \emph{how} provides us with a framework to structure those observations, and to predict and evaluate implications of the theory.

Language change can be studied at different scales. Phylogenetic approaches typically look across millennia \citep{bow18,greenhill2010shape} and concentrate on areas that are assumed to be stable. Other methods look at micro-levels of change; for example, \citet{Yao2018,hamilton2016diachronic} and \citet{eisenstein2014diffusion} study change at the range of decades and weeks respectively.


\subsection{Traditional explanations of language change}\label{seq:trad}
The current `received view’ of language change can be summarized as follows (necessarily with much loss of nuance; see further \citealt{hocjos96}). Language change begins with an innovation in a single language user. That innovation catches on and spreads through a  community, over time replacing older forms. Because not all members of a language community interact with each other all the time, innovations spread at different rates, and to different extents, across a language area. Thus dialects form, and those dialects eventually become sufficiently different that they come to be regarded as different languages. Innovations may also be introduced when speakers/signers of a language come into contact with a different language or dialect and adopt some of its features.

Most generative approaches to change assume that the point at which languages change is when children are acquiring language \citep{lig91,hal07}, a model that goes back ultimately to \citet{paul1880}. Yet we know that language acquisition is not the main driver of all language change. Language change in the historical record happens too fast for children to be solely involved.\footnote{Compare the arguments in \citet{darcy2017} for the recent spread of `like' as a discourse particle.} The evidence is overwhelming that children’s role is minimal \citep{ait03} in the spread of innovations. The errors that children make are not the main types of change we see in the record. Moreover, innovations are spread through social networks, and children acquiring language have peripheral positions in such networks.

The key questions model of change summarized in Figure~\ref{fig:keyqs}, though fairly common in evolutionary anthropology and in phylogenetic  approaches, is not the way historical linguistics has been conceptualized traditionally. \citet{WLH68} or \citet{lab01}, \citet{lig91}, and others in the generative tradition have often conceptualized the nature of the task of historical linguistics is being about the \emph{differences} between two stages of a language. That is a simpler problem, since it reduces language change to problem of edit distances. But it does not answer the questions we posed above, except inasmuch as identifying the differences -- that is, figuring out that something happened -- is just Stage 0 in understanding what happened, how it happened, and why. 


\subsection{Evolutionary views of change}
An alternative approach is a framework which treats language as a complex evolutionary system \citep[e.g.][]{bow18,mesoudi2011cultural,wedelexemplar2006}. This views language as a Darwinian system where changes are modeled through the key properties of variation, selection, and transmission.

In an evolutionary system, change is modeled as follows. The unit of study is the population; for language, our `population' could be a speech community or members of an ethno-linguistic group \citep{marlowe2005}. Such communities are inherently \emph{variable}: we know that not everyone speaks the same way, and that variation has social meaning. Systems which contain no variation cannot be modeled in an evolutionary framework. 

Much linguistic variation can be described in terms of social variables such as age, gender, socioeconomic class, geography, ethnicity, patri-group, moiety, and the like (though of course, not all of these variables  explain linguistic variation). Speakers do not use these variables deterministically, but with them index  aspects of social identity \citep{bucholtz2005identity}. Other inputs to the pool of variation include psychological and physiological aspects of language production and perception. For example, the fundamental frequency (or `pitch') of speech partly varies physiologically (taller people have deeper voices), partly socially (higher and lower pitch can index femininity and masculinity, respectively), and partly grammatically (for example, the difference between a declarative statement and a question can be signaled solely by an intonational rise at the end of the clause). 

Some of these variants are under \emph{selection} (positive or negative). Not all variants have equal chances of spreading within a community. Not all variants are under positive or negative selection; those that are are likely to change faster. Selection can be models as a set of bias biases in language transmission which inhibit or faciliate transmission. Such biases include acquisition, cognitive/physiological biases, and social biases.  

Over time, these biases affect the input that children are exposed to, as well as the ways adults use language. We see the results reflected over generations as ``change'' propagated through the linguistic record.

\begin{figure}[t]
    \centering
    \includegraphics[width=\linewidth]{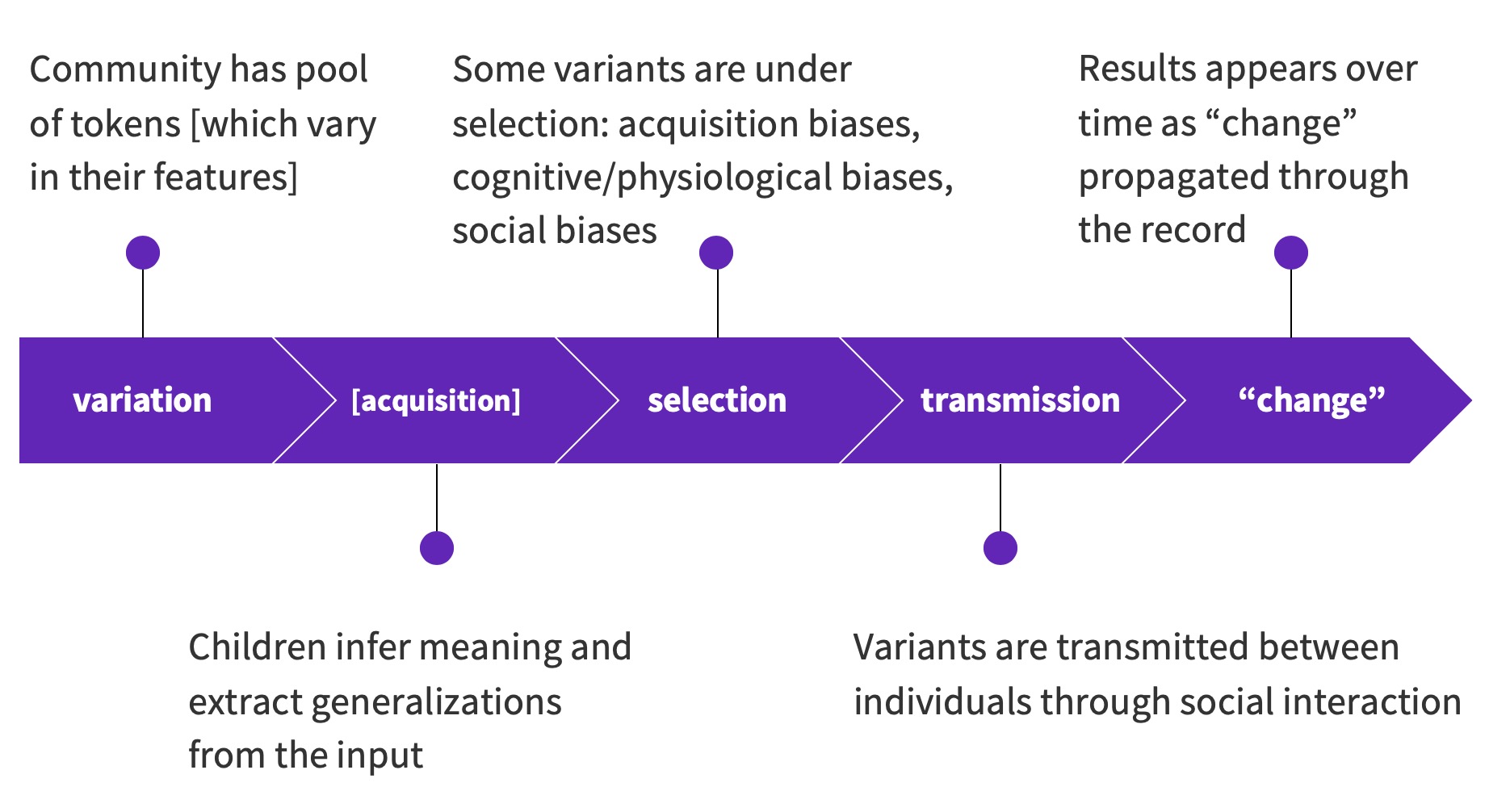}
    \caption{Schematic representation of language change in an evolutionary framework}
    \label{fig:change}
\end{figure}

Conceptualizing language change in this way has  consequences for how change is studied. Instead of looking across a system to extract generalizations, we are looking within a system for the points at which features vary. That is, we are not just comparing differences across points in time, but examining variation within a system and how that variation changes over time. Contrast semantic change studied by word embeddings, for example, where words are treated as discrete and uniform entities at each time point. As such, they are unable to distinguish between relative shifts in frequency of use among subsenses, and the spread of genuine innovations. The former may be a precursor to the latter, but the processes are not identical. 

Moreover, studying change in this way (correctly) entails that we not conceptualize change as `facilitating efficient communication'. This is a teleological view. Instead, biases and synchronic features of language make some changes more or less likely \citep[cf.][]{blevins2004}.

Finally, the transmission mechanism for language need not be strictly intergenerational. Taking an evolutionary view of language change does \emph{not} entail that it be studied with direct and concrete analogues to biological replication and speciation. A evolutionary view requires that there be a modeled transmission mechanism, not that the transmission mechanism exclusively involves transfer of material from parents to their children.



\section{Lexical replacement models}
\subsection{Types of lexical replacement}
With that background, let us now consider `change' specifically as applied to the lexicon. Like other parts of language, the lexicon is also constantly changing. The lexicon can be viewed as a set of mappings between forms, meanings, and the world. For example, the form we write as \emph{cat} maps to a concept, which relates to language users' knowledge of this animal in the real world. 

The following points summarize the types of lexical replacement that are possible in spoken and signed languages. Numerous works on semantic change have typologized the relationships between words and concepts at different stages in time \citep[cf.][]{tradas02}. Terms such as subjectification, meronymy, and amelioration all describe different relationships between words across time. Such points are, in this typology, all contained under the concept of ``semantic change''.

\begin{enumerate}
    \item Semantic change: that is, change in mappings between a lexical item, concepts, and world\vspace*{-3pt}
    \item Borrowing from other languages\vspace*{-3pt}
    \item Creation of words \emph{de novo}\vspace*{-3pt}
    \item (Loss)\vspace*{-3pt}
\end{enumerate}

As \citet{bender19} has noted, because of the heavy emphasis on English in NLP, the distinction between words and concepts is sometimes obscured. Yet it is vital when considering how concepts change. For example, if I describe a movement as \emph{catlike}, I am evoking aspects of the concept `cat', not a literal cat. (Someone can walk in a catlike fashion without, for example, being furry or having a tail.) 

An emphasis on typologically similar and closely related languages is also problematic for studying tendencies. For example, \citet{hamilton2016cultural} argue as an absolute that nouns are more likely to undergo irregular cultural shifts (e.g. expansion due to technological innovations) while verbs are more likely to show regular processes of change, such as drift. Such a view does not take into account that verb numbers differ extensively across languages, and the functional load, levels of polysemy, and lexicalization patterns for events also differ -- points that \citet{hamilton2016diachronic} showed were important in assessing likelihood of change. Technological innovation, while exceptionally salient to those who work in NLP, is unlikely to have been the same driving factor in semantic change across most of human history. And indeed, it plays a small role in the literature on lexical replacement, where euphemism, metaphorical extension, and bleaching play more important roles.

A further type of lexical replacement involves borrowing \citep{hastad09}. Both borrowing and creation of words from new resources involve the innovation of mappings between words and concepts within a linguistic system. In the former, lexical material is adapted from another language, while in the latter, it is created from language-internal resources or innovated from scratch. Languages differ in the extent to which novel word formation is utilized, and the strategies, from compounding to acronyms to blends, also vary greatly. Furthermore, there is variation in the extent to which language users borrow words (see further \citealt{bow11plos}), but there are regularities in which words are more likely to be borrowed. Word creation has played a role in NLP approaches to semantic change because of the focus on named entity identification, but it is a small part of change overall.

\subsection{Evolutionary semantic change}\label{sec:alternatives}
Such changes can be modeled in an evolutionary framework. Some variation is neutral (not under selection). For example, speakers of American English have several distinct systems of contrast in the meanings of the words `cobweb' and `spiderweb':\footnote{The source of this observation is 4 years of polling historical linguistics students at Yale.}

\begin{itemize}
    \item The two words are synonymous;
    \item Spiderwebs are spiral or wheel-shaped, cobwebs are collapsed;
    \item Spiderwebs have spiders in them, other items are cobwebs (including abandoned but intact wheel webs);
    \item Spiderwebs have spiders, while cobwebs are synonymous with dirt or dust bunnies (detritus that is cleaned when cleaning a house). That is, cobwebs are not necessarily old spiderwebs but could be from other material.
\end{itemize}

\noindent Speakers are unaware of these differences in semantic distinctions, and the variants do not clearly pattern by age, gender, or geography. Such variation is not under selection and is below the level of consciousness. It is, however, very hard to detect (not least because it is usually also invisible to researchers).

Other selectional pressures skew change. Such biases include (but are not limited to) meaning transmission failure and speaker attitudes. For example, there is a bias against using words with novel denotations. Meaning is conventionalized, which is what prevents English speakers from calling a `{\Large$\star$}' a `sun'. However, language users do make creative and novel associations between objects, which do over time end up as change. For example, several Pama-Nyungan subgroups have words for `eye' which are etymologically connected to `seeds' (compare Wati and Pama-Maric languages, which have independently shifted \emph{*kuru} `seed' to `eye'; the Yolngu language Yan-nha{\engma}u has a single term \emph{ma{\engma}utji}, which means `eye', `well', and `seed'. To study such changes, it is vital to have a good empirical basis for the possibilities for polysemy and shift. \citet{list2013using} provides an example using translation equivalents across languages from different families. 





Finally, words can also fall out of use. They may be tabooed through necronym replacement or protective euphemism, or lost when the knowledge of the concepts they represent is also lost (such as ethnobiological knowledge in many urban English speakers).  

In summary, semantic change can be modeled in an evolutionary framework, where meanings vary, have positive or negative selectional biases, and are transmitted through language use. If a word is not used, it is not transmitted. Such a view provides a clue to \citeauthor{hamilton2016diachronic}'s findings about polysemy and and frequency. Words are more likely to change if they have low frequency, because speakers have less information about meaning, making them more vulnerable to reinterpretation or replacement (further eroding their frequency). Words are also more likely to change if they exhibit high polysemy, perhaps because they are both more ambiguous and more likely to be further extended. 

\subsection{Word embeddings}\label{sec:embeddings}
With this theoretical background, let us now turn to an evaluation of methods. Word embeddings \citep{turney2010,Kulkarni2015} are an increasingly common tool for studying change in vocabulary over time. They rely on the intuition that ``you can know a word by the company it keeps'' \citep[11]{firth1957}, and by studying the changes in word use it is possible to quantify and further study language change. 

Critiques of the effectiveness of using word embeddings to study change are well known. \citet{dubossarsky2017outta} and \citet{tahmasebi2018survey} have pointed out issues that limit the utility of embeddings for studying change, such as the necessity for large corpora, the brittleness of results, and the lack of ability to study word senses independently. This latter point is particularly important for theories of meaning change, since as argued above, understanding variation is a prerequisite to an adequate modeling of the evolution of linguistic systems over time.

Embeddings across massive corpora assume that all speakers have the same knowledge of the vocabulary of their language. That is simply not true, as illustrated by the simple example in~\S\ref{sec:alternatives} above. Not all speakers/signers know all the words of their languages. Using embeddings across many speakers and documents also conflates real-world knowledge (e.g. Linnaean classification) with linguistic knowledge. For example, I do not need to know that a koala is a member of the genus Phascolarctos to know what a koala is, any more than the etymology (from Daruk \emph{kula}) is part of the meaning. Yet because word embedding models use encyclopedic corpora such as Wikipedia, they tend to be skewed towards such information.



Finally, embedding changes conflate changes in frequency of a word with conceptual changes, further obscuring mechanisms of change.  \citet{Yao2018} identify shifts in frequency and use this as a diagnostic for language change. They use the example of `\emph{apple}'s vectorization changing over time from being more similar to other fruit to being more similar to computer equipment and software. However, just because apple is now more associated in their corpus with software than with fruit, it doesn't entail that the meaning of the word has actually changed over that time period. It is a possible precursor to a change where a word goes through a period of variation and polysemy (an A, A$\sim$B, B change), but that is not the only type of change. For a similar problem, see \citet{Kulkarni2015} on word usage time series, and for a more nuanced view, \citealt{DBLP}. If we are to study change, we can't just abstract away from variation in the data as ``noise''. Variation leads to change, and not all differences are changes.







\section{Lexical replacement and phylogenetics}
\subsection{Stability and meaning}
So far, I have concentrated discussion on variation and change. However, for studying change at the macro-level, across phylogenetic time, we require  items which have high semantic stability. Evolutionary approaches to language split use lexical replacement to model language evolution. That is, they take presumed stable (but nonetheless varying) meaning categories and use the variation in the realization of those meanings to build a model of language split, from which the phylogeny is recovered. Such work is now well established in the literature on language change and the reader is referred to \citet{dunn2014} and \citet{bow18} for summaries. State of the art methods use Bayesian inference; see \citet{bouckaert2018} for explanation and details of priors, cognate models, and data treatment.

Such methods can be used to study semantic change over a phylogeny. They are particularly useful for studying the lexicalization of oppositions within a small semantic space. For example, \citet{haybow16} used such methods to see how color terms changed across the Australian family Pama-Nyungan. The visible color spectrum is modeled as partitioned by vocabulary \citep{rkc2005}. These partitions obey evolutionary principles. There is variation (people don't have full agreement in the assignment of lexicon to the visible spectrum, and color terms vary across languages); transmission (color terms are acquired and transmitted with other aspects of language) and selection (there are physiological constraints on perception (which are also variable), for example, and visual exemplars which tend to lexicalize as color terms; cf.\ `orange'). Keeping the conceptual space constant and varying the partitioning avoids the problem that other types of change are happening simultaneously. That is, we can't study the evolution of particular words in many domains because the words fall out of use or are replaced too many times across the tree. 


These models require cognate evolution models. Currently, the main one is Brownian Motion (that is, random change across a tree). Such models fit these types of change well, and allow us to evaluate the effectiveness of such models as well as probabilistically reconstructing ancestral states.

\subsection{Lexical replacement in phylogenetics}\label{sec:repl}
A final illustration of evolutionary methods for meaning change and lexical replacement concerns a practical issue for phylogenetics:  the `legacy problem' of Swadesh wordlists. Since \cite{swadesh1952,swadesh1955}, linguists have been using similar lists of so-called `basic vocabulary' to construct cognate evolutionary matrices.\footnote{A `cognate' is a a word which shares an evolutionary history of descent with other words. English `fish' and German `Fisch' are cognate, because they continue the same form-meaning correspondence from an ancestor language. English `much' and Spanish `mucho' are not cognate, despite their similarity in form, because they continue different lexical roots. `Much' continues Old English \emph{mickel} (ultimately from an Indo-European root meaning `big, great', while Spanish \emph{mucho} continues Latin \emph{multus} `much, many', ultimately from a root meaning `crumpled'. } These wordlists are now a sample of convenience, as lexical resource collection has prioritized vocabulary from Swadesh lists. Other work \citep{mcmmcm06} has reduced the number of comparison items even further. \citet{rama2018towards} estimate the number of items needed for small phylogenies; however, they do not take stability  into account. Their estimate concludes approximately 30 data points per language in the phylogeny. However, the number of such data points varies with both the number of meaning classes and their stability. To illustrate for 300 Pama-Nyungan languages, the number of cognates per meaning class in the Swadesh 200 list ranges from 40 to 199, and the number of languages with a singleton cognate in a meaning class ranges from 20 (for the second person plural pronoun) to 126 (for translations of the concept `small').

\begin{figure}[t]
    \centering
    \includegraphics[width=\linewidth]{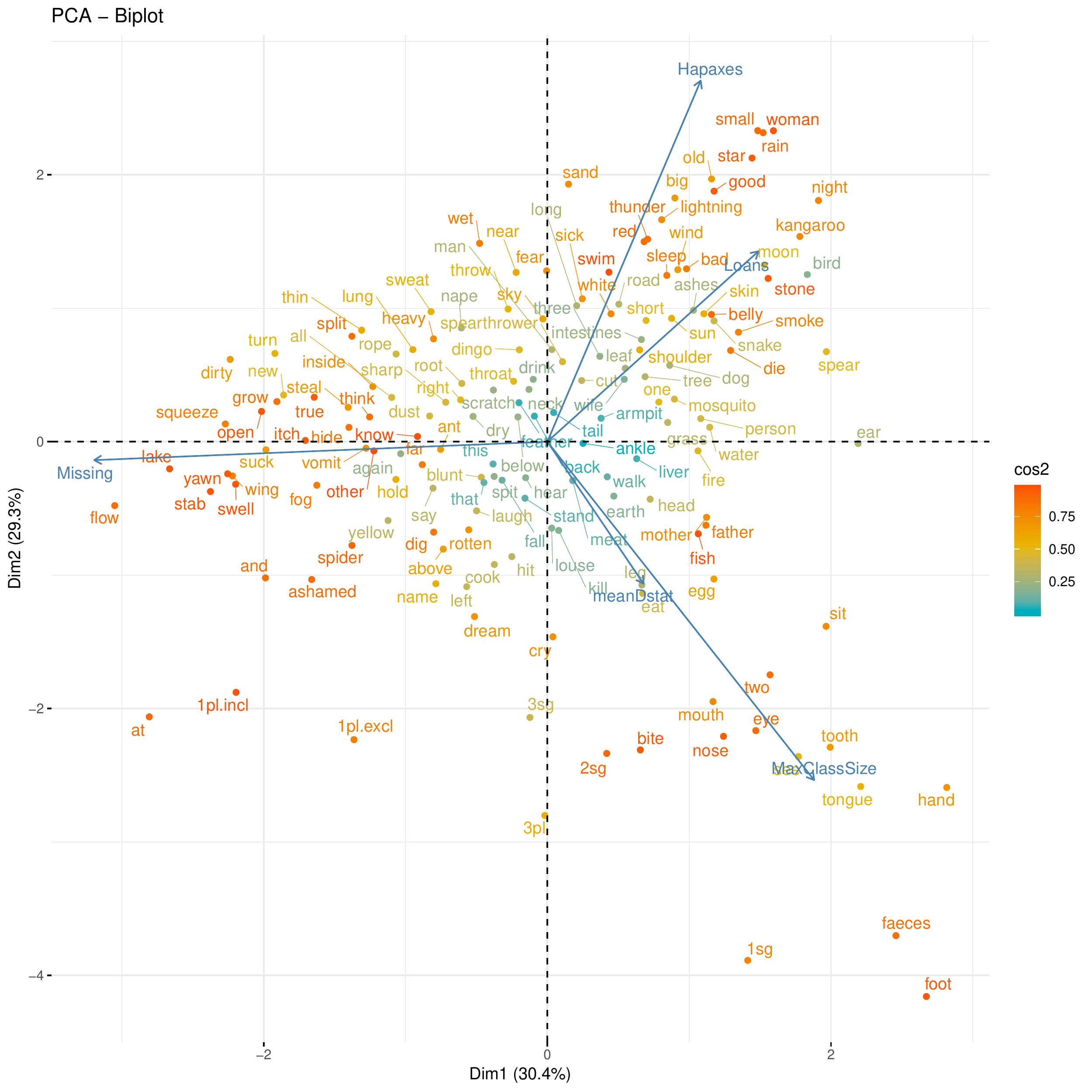}
    \caption{PCA and loadings for meaning classes}
    \label{fig:pamPCA}
\end{figure}

The effect of the choice of vocabulary on phylogeny is not well studied. \citet{bouckaert2018} point out that the difference between \citet{bowatk12} and their phylogeny includes additional words; using an additional 20 vocabulary items changed the classification of some languages to be more in line with established subgrouping based on grammatical features. We know that loan rates affect suitability for phylogenetic inference, and that loan rates in basic vocabulary vary. We are left with Swadesh lists being the default instrument for inference, yet they are based on a list whose membership was determined by, to put it bluntly, one person's suggestion of what might be useful to diagnose remote relationships 70 years ago, not on a principled decision of stability in meaning classes.

Many factors contribute to go into making a meaning class a good or poor choice for phylogenetics. If the meaning class is too stable, there is not sufficient variation to recover and date phylogenetic splits. If a word is widely loaned, that will make the evolutionary history harder to uncover and reduce phylogenetic signal. If an item changes too fast, or there are too many singleton reflexes, there is less informative signal higher in the tree. Homoplasy (convergent evolution) is also problematic, as it it difficult to detect and can lead to false language groupings. 
In order to evaluate the suitability of individual meaning classes, I coded cognate sets in the material used in \citet{bouckaert2018} and \citet{bow11plos} for number of loan events, informativeness of phylogenetic signal (D statistic; see \citealt{fritz2010selectivity}), number of singletons, amount of missing data, and mean and maximum meaning class size (that is, how many languages attest a particular cognate in that meaning class). Figure~\ref{fig:pamPCA} plots the first two PCA and clusters meaning classes based on these variables, using the fviz\underline{ }cluster() function in the Factoextra package in R \citep{facto}. 
The largest factor contributing to dimension 1 is how much data is missing, while dimension 2's largest contribution is the number of singleton cognates per meaning class. Meaning classes which score relatively highly on dimension 1 and relatively low on dimension 2 are most likely to be optimal for phylogenetic analysis. However, items solely taken from the southeast quadrant are the most stable, and therefore likely to lead to underestimates of splits.

\section{Conclusion}
In conclusion, evolutionary approaches to language change provide explicit ways of modeling semantic shifts and lexical replacement. They provide researchers with a structure for examining the facts of language differences, the mode and tempo  of language change, and a way of framing questions to lead to an understanding of why languages change the way they do. In all this, however, variation is key -- it provides the seeds of change,  allows the identification of change in progress, and the absence of variation makes it possible to study stability and shift across millennia.


\bibliography{acl2019}
\bibliographystyle{acl_natbib}

\end{document}